\documentclass[letterpaper]{article}
\usepackage{aaai}
\usepackage{times}
\usepackage{helvet}
\usepackage{courier}
\usepackage{bm}
\usepackage{amssymb}
\usepackage{graphicx}
\usepackage{wrapfig}
\usepackage{amsmath}
\usepackage{url}
\usepackage{xcolor}
\usepackage{caption}
\usepackage{comment}
\usepackage[justification=centering]{caption}
\usepackage{soul}
\usepackage{subcaption}

\begin{document}

\title{Quantifying Classification Uncertainty using Regularized Evidential Neural Networks}

\author{Xujiang Zhao, Yuzhe Ou\\
{UT Dallas, TX 75080}\\
\{xujiang.zhao, yxo170030\}\\
@utdallas.edu
\And
Lance Kaplan\\
{US Army Research Lab}\\ Adelphi, MD 20783, USA\\
lkaplan@ieee.org
\And
Feng Chen\\
{UT Dallas, TX 75080}\\
fchen5@albany.edu\
\And
Jin-Hee Cho\\
{Virginia Tech}\\
Falls Church, VA 22043\\
jicho@vt.edu
}
\maketitle
\begin{abstract}

Traditional deep neural nets (NNs) have shown the state-of-the-art performance in the task of classification in various applications. However, NNs have not considered any types of uncertainty associated with the class probabilities
to minimize risk due to misclassification under uncertainty in real life. Unlike Bayesian neural nets indirectly infering uncertainty through weight uncertainties, evidential neural networks (ENNs) have been recently proposed to support explicit modeling of the uncertainty of class probabilities.  It treats predictions of an NN as subjective opinions and learns the function by collecting the evidence leading to these opinions by a deterministic NN from data. However, an ENN is trained as a black box without explicitly considering different types of inherent data uncertainty, such as vacuity (uncertainty due to a lack of evidence) or dissonance (uncertainty due to conflicting evidence). This paper presents a new approach, called a {\em regularized ENN}, that learns an ENN based on regularizations related to different characteristics of inherent data uncertainty. Via the experiments with both synthetic and real-world datasets, we demonstrate that the proposed regularized ENN can better learn of an ENN modeling different types of uncertainty in the class probabilities for classification tasks. 
\end{abstract}

\section{Introduction} \label{sec:intro}
Inherent data uncertainty due to different root causes has been serious hurdles in providing effective solutions for real world problems.  Critical safety issues often may occur because different types of uncertainty have not been sufficiently considered and even some perceived uncertainty tends to be improperly interpreted because of lack of knowledge on the cause of the uncertainty.  For example, when an autonomous vehicle misclassifies objects, this may lead to highly risky situations.  Deep learning (DL) models have recently gained tremendous attention in the data science community~\cite{kipf2016semi,velickovic2018graph}.  Despite their superior performance in various tasks (e.g., classification or regression), they are limited in dealing with different types of data uncertainty.  Predictive uncertainty estimation using Bayesian NNs (BNNs) has been explored for classification prediction or regression in the computer vision applications~\cite{malinin2018predictive}. They considered well-known uncertainty types, such as aleatoric uncertainty and epistemic uncertainty, where aleatoric uncertainty only considers data uncertainty caused by statistical randomness (e.g., observation noises) while epistemic uncertainty refers to model uncertainty introduced by limited knowledge or ignorance in collected data.  On the other hand, in the belief / evidence theory, Subjective Logic (SL)~\cite{josang2018uncertainty} considered vacuity (or a lack of evidence) as the key dimension of uncertainty. A recent work on SL has defined other aspects of uncertainty, such as dissonance, consonance, vagueness, and monosonance~\cite{josang2018uncertainty}. 

Traditional deep neural nets (NNs) have been popularly applied in solving classification tasks. However, NNs have not been used to deal with uncertainties associated with the class probabilities to minimize risk due to misclassification under uncertainty in real life applications. Methods using BNNs have been proposed to indirectly predict uncertainty through weight uncertainties. Recently, techniques using evidential neural networks (ENNs) have been proposed to support explicit modeling of the uncertainty of class probabilities. It treats predictions of a neural net as subjective opinions and learns the function that collects the evidence leading to these opinions by a deterministic NN from data. However, an ENN is still trained as a black box without explicitly considering different types of data uncertainty (e.g., vacuity or dissonance).  This paper presents a new approach that learns an ENN based on regularizations related to different types of inherent data uncertainty, which we call it {\em regularized ENNs}.

Our {\bf key contributions} are summarized as follows:
\vspace{-1mm}
\begin{enumerate}
\item Our proposed regularized ENNs-based method is the first that explicitly considers different types of inherent data uncertainty in learning an ENN for quantifying uncertainty in classification tasks.  
\vspace{-2mm}
\item We validated the performance of our regularized ENNs-based method via extensive simulation experiments using both synthetic and real-world datasets.  We conducted the performance comparison of predicting different types of uncertainty when applying classification methods based on the regularized ENNs, ENNs (without regularization), and NNs. 
\vspace{-1mm}
\end{enumerate}

\section{Motivational Scenarios} \label{sec:motivation}
This paper focuses on scenarios where decision making is modeled as a classification problem withe the aim of quantifying uncertainty of decision making in terms of classification prediction. Such cases is common in the government and public sectors with the following examples:
\begin{itemize}
    \item \textit{Self-driving cars safety}: An autonomous car may deploy a deep learning system that fails to distinguish the white side of a truck against bright sky because such type of the truck has not been known before. In this case, vacuity (i.e., a lack of evidence) can represent a measure of the extent the detected object belongs to a class of cars that has never been known before (i.e., not in a considered domain).  Further, if the truck is not detected as a single class type, it increases uncertainty due to vagueness. Or if the tuck is detected with either one type or the other due to conflicting evidence, it also increases uncertainty because of conflicting evidence.  
    
    \item \textit{Public health}: The outbreak of an unknown disease occurs and we observe that the correlated symptoms and the spread patterns cannot be reliably predicted. As described in other examples, a lack of evidence (i.e., vacuity), conflicting evidence supporting either one or the other (i.e., dissonance among evidence) or vagueness by failing in identifying as a single object may significantly lead to increasing uncertainty which introduces high challenging in decision making.
    
    \item \textit{Security in network infrastructure}: A power network system can be targeted by a cyberattack, which can break down the system which can introduce tremendous damage because this attack is not timely detected.  Uncertainty introduced by vacuity, dissonance, or vagueness (as each term is discussed above) can delay decisions that can immediately take actions to minimize the damage.  
\end{itemize}
To focus subsequent discussion on a single concrete example, we will consider image classification, a critical component in autonomous cars. We will quantify the uncertainty of an image classifier in recognizing objects of a known or unknown class in images using neural network models.  
\section{Related Work} \label{sec:related-work}

{\bf Uncertainty Quantification in Bayesian Deep Learning (BDL)}: Machine/deep learning (ML/DL) researchers considered {\it aleatoric} uncertainty (AU) and {\it epistemic} uncertainty (EU) based on BNNs for computer vision applications.  AU consists of homoscedastic uncertainty (i.e., constant errors for different inputs) and heteroscedastic uncertainty (i.e., different errors for different inputs)~\cite{gal2016uncertainty}. A BDL framework was presented to estimate both AU and EU simultaneously in regression settings (e.g., depth regression) and classification settings (e.g., semantic segmentation)~\cite{kendall2017uncertainties}.  Later {\em distributional uncertainty} is defined based on distributional mismatch between the test and training data distributions~\cite{malinin2018predictive}.  {\em Dropout variational inference}~\cite{gal2016dropout} was proposed as one of key approximate inference techniques in BNNs.  Other methods~\cite{eswaran2017power,zhang2018bayesian} measured overall uncertainty in node classification but were not focused on uncertainty decomposition and GNNs. 

\vspace{1mm}
\noindent {\bf Uncertainty Quantification in Belief/Evidence Theory}: In belief/evidence theory domain, uncertainty reasoning has been substantially explored in Fuzzy Logic~\cite{de2018intelligent}, Dempster-Shafer Theory (DST)~\cite{sentz2002combination}, or Subjective Logic (SL)~\cite{josang2016subjective}.  Unlike the efforts in ML/DL above, belief/evidence theory focused on reasoning of inherent uncertainty in information resulting from unreliable, incomplete, deceptive, and/or conflicting evidence.  SL considered uncertainty in subjective opinions in terms of {\em vacuity} (i.e., a lack of evidence) and {\em vagueness} (i.e., failure of discriminating a belief state)~\cite{josang2016subjective}. Recently, other dimensions of uncertainty have been studied, such as {\em dissonance} (due to conflicting evidence) and {\em consonance} (due to evidence about composite subsets of state values)~\cite{josang2018uncertainty}.

\section{Background} \label{sec:background}
This section provides the background knowledge to understand this work including: (1) subjective opinions in SL; (2) uncertainty dimensions in a subjective opinion; and (3) evidential NNs to predict subjective opinions. 

\vspace{1mm}

\noindent {\bf Subjective Opinions in SL}: A multinomial opinion in a given proposition $x$ is represented by $\omega_Y = (\bm{b}_Y, u_Y, \bm{a}_Y)$ where a domain is $\mathbb{Y} \equiv \{1, \cdots, K\}$ and the additivity requirement of $\omega_y$ is given as $\sum_{y \in \mathbb{Y}} \bm{b}_Y(y) + u_Y = 1$.  To be specific, each parameter indicates,
\begin{itemize}
\item $\bm{b}_Y$: {\em belief mass distribution} over $\mathbb{Y}$;
\item $u_Y$: {\em uncertainty mass} representing {\em vacuity of evidence};
\item $\bm{a}_Y$: {\em base rate distribution} over $\mathbb{Y}$.
\end{itemize}

The projected probability distribution of a multinomial opinion is given by:
\begin{equation} \label{eq:multinomial-projected}
\mathbf{P}_Y(y) = \bm{b}_Y(y) + \bm{a}_Y(y) u_Y,\;\;\; \forall y \in \mathbb{Y}
\end{equation}  

Multinomial probability density over a domain of cardinality $K$ is represented by the $K$-dimensional Dirichlet PDF where the special case with $K=2$ is the Beta PDF as a binomial opinion. Denote a domain of $K$ mutually disjoint elements in $\mathbb{Y}$ and $\alpha_Y$ the strength vector over $y \in \mathbb{Y}$ and ${\bf p}_Y$ the probability distribution over $\mathbb{Y}$. Dirichlet PDF with ${\bf p}_Y$ as $K$-dimensional variables is defined by:
\begin{eqnarray} \label{eq:multinomial-dir}
\mathrm{Dir}(\bm{p}_Y; {\bm \alpha}_Y) = \frac{1}{B({\bm \alpha}_Y)} \prod_{y \in \mathbb{Y}} \bm{p}_Y (y) ^{({\bm \alpha}_Y(y)-1)}
\end{eqnarray} 
where $\frac{1}{B({\bm \alpha}_Y)} = \frac{\Gamma \Big(\sum_{y \in \mathbb{Y}} {\bm \alpha}_Y (y)\Big)}{\prod_{y \in \mathbb{Y}} ({\bm \alpha}_Y (y))}$, ${\bm \alpha}_Y(y) \geq 0$, and ${\bf p}_Y (y) \neq 0$ if ${\bm \alpha}_Y (y) < 1$.

We term \textit{evidence} as a measure of the amount of supporting observations collected from data in favor of a sample to be classified into a certain class. Let ${\bf r}_Y(y) \ge 0 $ be the evidence derived for the singleton $y\in \mathbb{Y}$.  The total strength ${\bm \alpha}_Y(y)$ for the  belief of each singleton $y \in \mathbb{Y}$ is given by: 
\begin{eqnarray} \label{eq:multinomial-alpha}
{\bm \alpha}_Y(y) = \bm{r}_Y(y) + \bm{a}_Y(y) W, \\
\text{where } \bm{r}_Y(y) \geq 0, \; \; \forall_y \in \mathbb{Y} \nonumber
\end{eqnarray}
where $W$ is a non-informative weight representing the amount of uncertain evidence and $\bm{a}_Y(y)$ is the base rate distribution.  Given the Dirichlet PDF, the expected probability distribution over $\mathbb{Y}$ is:
\begin{equation} \label{eq:multinomial-expected}
\small
\mathbb{E}_Y(y) = \frac{{\bm \alpha}_Y (y)}{\sum_{y_i \in \mathbb{Y}} {\bm \alpha}_Y (y_i)} = \frac{\bm{r}_Y(y) + \bm{a}_Y(y) W}{W + \sum_{y_i \in \mathbb{Y}} \bm{r}_Y(y_i)}
\end{equation}
The observed evidence in the Dirichlet PDF can be mapped to the multinomial opinions by:

\begin{equation} \label{eq:multinomial-belief}
\small
\bm{b}_Y(y) = \frac{\bm{r}(y)}{S}, \;
u_X = \frac{W}{S},  
\end{equation}
where $S = \sum_{y_i \in \mathbb{Y}} {\bm \alpha}_Y(y_i)$. 
With loss of generality, we set  $\bm{a}_Y(y) = \frac{1}{K}$ and the non-informative prior weight (i.e., $W = K$), and hence $\bm{a}_Y(y)\cdot  W = 1$ for each $y \in \mathbb{Y}$.

\begin{figure*}
\centering
\begin{subfigure}{.33\textwidth}
  \includegraphics[width=.90\linewidth]{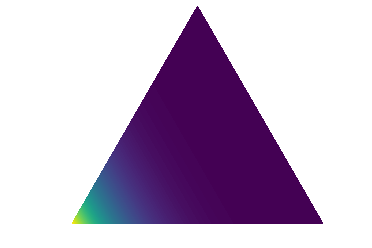}
  \captionof{figure}{ $\alpha=[6,1,1], \mathrm{Vac}=0.375$, \\ $\mathrm{Diss}=0$}
  \label{fig:Dirichlet1}
\end{subfigure}
\centering
\begin{subfigure}{.33\textwidth}
  \includegraphics[width=.90\linewidth]{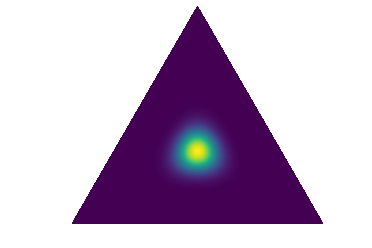}
  \captionof{figure}{ $\alpha=[20,20,20], \mathrm{Vac}=0.05$, \\ $\mathrm{Diss}=0.95$}
  \label{fig:Dirichlet2}
\end{subfigure}
\begin{subfigure}{.33\textwidth}
  \includegraphics[width=.90\linewidth]{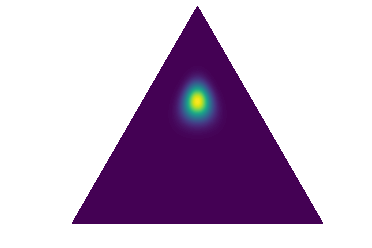}
  \captionof{figure}{$\alpha=[20,20,50], \mathrm{Vac}=0.33$, \\ $\mathrm{Diss}=0.592$}
  \label{fig:Dirichlet3}
\end{subfigure}
\caption{Illustration of different vacuity and dissonance of subjective opinions based on their evidence measures. }
\label{fig:Dirichlet}
\end{figure*}

\vspace{1mm}
\noindent {\bf Uncertainty Characteristics of Subjective Opinions}: In~\cite{josang2018uncertainty}, the multidimensional uncertainty dimensions of a subjective opinion based on the formalism of SL are discussed, such as singularity, vagueness, vacuity, dissonance, consonance
and monosonance.  These different dimensions of uncertainty can be observed from binomial, multinomial, or hyper opinions depending on their characteristics (e.g., vagueness is only observed in hyper opinions when a composite belief exists). As we deal with a multinomial opinion in this work, we discuss two main types of uncertainty dimensions that can be estimated in a multinomial opinion, which are {\em vacuity} and {\em dissonance}. 
 
The main cause of vacuity is due to a lack of evidence or knowledge, which corresponds to uncertainty mass, $u$, of an opinion in SL as:
\begin{eqnarray}
\text{Vac}({\bm \alpha}_Y) = \frac{W}{S}. \label{vacuity}
\end{eqnarray}
This type of uncertainty refers to uncertainty because the analyst does not have sufficient information or knowledge to analyze the uncertainty.

The {\em dissonance} of an opinion can be caused by the same amount of conflicting evidence and is estimated by the difference between singleton belief masses (e.g., class labels), leading to `inconclusiveness' in decision making situations. For example, given a four-state multinomial opinion with $(b_1, b_2, b_3, b_4, u, a) = (0.25, 0.25, 0.25, 0.25, 0.0, a)$ based on Eq.~\eqref{eq:multinomial-belief}, although $u$ (vacuity) is zero, one cannot make a decision if there are the same amounts of beliefs supporting respective beliefs.  Given a multinomial opinion with non-zero belief masses, the measure of dissonance can be obtained by:
\begin{equation}
\label{eq:belief-dissonance-multi}
\text{Diss}({\bm \alpha}_Y) = \sum\limits_{y_{i}\in \mathbb{Y}}\left(\frac{\bm{b}_{Y}(y_{i})\!\!\!\! \sum\limits_{y_j \in \mathbb{Y}\setminus y_i}\!\!\!\!\!\bm{b}_{Y}(y_{j}) \mbox{Bal}(y_{j},y_{i})}{\sum\limits_{y_j \in \mathbb{Y}\setminus y_i}\bm{b}_{Y}(y_{j})}  \right),
\end{equation}
where the relative mass balance between a pair of belief masses $\bm{b}_{Y}(y_{j})$ and $\bm{b}_{Y}(y_{i})$ is expressed by:
\begin{equation}
\label{eq:belief-balance}
\mbox{Bal}(y_{j},y_{i}) = 1 - \frac{|\bm{b}_{Y}(y_{j})-\bm{b}_{Y}(y_{i})|}{\bm{b}_{Y}(y_{j})+\bm{b}_{Y}(y_{i})}, 
\end{equation}
Recall that we measure the dissonance only when the belief mass is non-zero. If all belief masses are zero with vacuity being 1 (i.e., $u_Y=1$), the dissonance (or balance) will be set to zero.

The relative mass balance has its maximum at 1 when $\bm{b}_{Y}(y_{j}) = \bm{b}_{Y}(y_{i})$. The relative mass balance has the minimum at 0 when one of the belief masses equals 0.  In case of the zero sum of belief masses in both the nominators and the denominators of Eq.~\eqref{eq:belief-dissonance-multi} and Eq.~\eqref{eq:belief-balance}, it must be assumed that the fraction of the sums of belief masses equals 1.  But notice that even with high vacuity (a lack of evidence), high dissonance can be observed.

The above two uncertainty measures (i.e., vacuity and dissonance) can be interpreted using class-level evidence measures of subjective opinions. As in Figure~\ref{fig:Dirichlet}, given three classes (1, 2, and 3), we have three subjective opinions $\{{\bm \alpha}_1, {\bm \alpha}_2, {\bm \alpha}_3\}$, represented by the three-class evidence measures as: ${\bm \alpha}_1 = (6, 1, 1), {\bm \alpha}_2 = (20, 20, 20), {\bm \alpha}_3 = (20, 20, 50)$. The vacuity and dissonance measures of the three subjective opinions can be calculated via Eq.~\eqref{vacuity} and Eq.~\eqref{eq:belief-dissonance-multi}, respectively. 

When the evidence measures of the three classes of a subjective opinion decrease, the corresponding vacuity increases. When the evidence measures of the three classes of a subjective opinion become closer (and therefore are more conflicting with each other), the corresponding dissonance increases.

\vspace{1mm}
\noindent {\bf Evidential Neural Networks (ENNs)}:
The gold standard for DNNs is to use the softmax operator to convert the continuous
activations of the output layer to class probabilities. Although the softmax function provides a point estimate for the class probabilities of a sample, it does
not provide the associated uncertainty. ENNs are similar to classical NNs, except that the softmax layer is replaced by an activation layer (e.g., ReLU) to ascertain
non-negative output, which is taken as the evidence vector for the predicted Dirichlet distribution. 

Given a sample $i$, let $f({\bf x}_i | \Theta)$ represent the evidence vector predicted by the network for the classification, where ${\bf x}_i \in \mathbb{R}^L$ is the observation of a feature in this sample and $\Theta$ is a set of network parameters. Then, the corresponding Dirichlet distribution has
parameters ${\bm \alpha}_i = f({\bf x}_{i} | \Theta) + 1$. Once the parameters of this distribution are calculated, its mean (i.e.,
${\bm \alpha}_i / S_i$) can be taken as an estimate of the class probabilities.

Let ${\bf y}_i$ be a dummy vector encoding the ground-truth class of observation ${\bf x}_i$ with $y_{ij}=1$ and $y_{ik} = 0$
for all $k \ne j$, and $\alpha_i$ be the parameters of the Dirichlet density on the predictors. The Dirichlet density $\text{Dir}({\bf p}_i ; {\bm \alpha})$ is the prior on the Multinomial distribution $\text{Multi}({\bf y}_i | {\bf p}_i)$. The following sum of squared loss (SSL) is used to estimate the parameters ${\bm \alpha}_i$ based on the sample $i$: 

\begin{eqnarray}
\mathcal{L}(f({\bf x}_i|\Theta), {\bf y}_i) =\int  \frac{\|{\bf y}_i - {\bf p}_i\|_2^2}{B({\bm \alpha}_i)} \prod_{j =1}^K p_{ij} ^{(\alpha_{ij}-1)} d{\bf p}_i \nonumber \\=\sum_{j=1}^K \Big(y_{ij}^2 - 2 y_{ij}\mathbb{E}[p_{ij}] + \mathbb{E}[p_{ij}^2]\Big). 
\end{eqnarray}

In the current design of ENNs, an ENN is trained as a black box using the above squared error function as the loss function with a regularization term to promote epistemic uncertainty 
\begin{equation}
\begin{split}
    \mathcal{L}(\Theta) = \mathbb{E}&_{(\mathbf{x}_i,\mathbf{y}_i)\sim\mathcal{D}}\left[ \mathcal{L}({\bm \alpha}_i,{\bf y}_i) \right .\\
    & + \left . \lambda_t KL\left[ \text{Dir}(\mathbf{p}_i;{\bm \alpha}^{-y_i}_i)||\text{Dir}(\mathbf{p}_i;\mathbf{1})\right] \right],
\end{split}
\end{equation}
where $KL[\cdot]$ represent the Kullback-Leibler divergence.
Note that ${\bm \alpha}_i^{-y_i}$ is the evidence predicted by the network except that the ground truth class evidence is set to zero.  The regularization term is necessary to bias the network to generate high epistemic uncertainty where it is likely to make errors such as decision boundaries. However, the training of ENN does not necessarily bias it to return high epistemic far from the training data at points also far from decision boundaries. 

Given the various characteristics of uncertainty that subjective opinions can describe, we propose a regularized ENN in the section below to learn an NN that is more effective to quantify uncertainty for classification tasks through the regularization on multidimensional uncertainty, such as vacuity and dissonance. 
\begin{figure}[th]
\vspace{-3mm}
\centering
\includegraphics[width=0.45\textwidth]{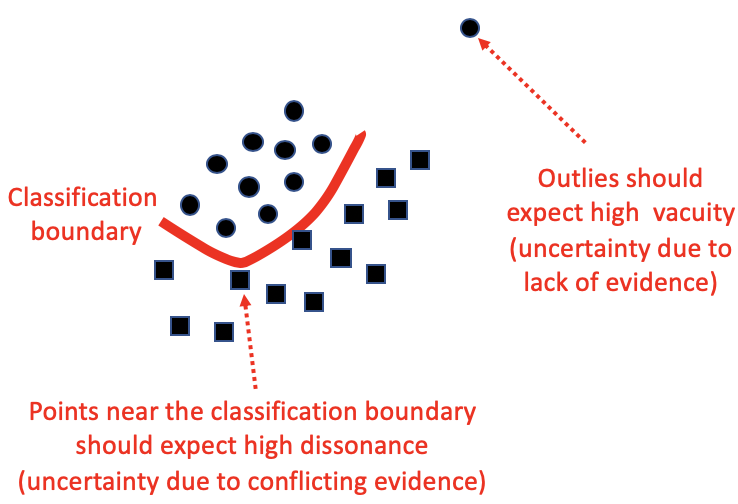}
\centering
\caption{Illustration of boundary points with high dissonance and outliers (or OOD samples) with high vacuity. }\label{fig1: example}
\vspace{-5mm}
\end{figure}

\vspace{1mm}
\noindent {\bf Regularized ENNs}: Given a set of samples  $\mathcal{D} = \{({\bf x}_1, {\bf y}_1), \cdots, ({\bf x}_N, {\bf y}_N)\}$, we can identify some samples that should expect high vacuity or dissonance using heuristic ways. As shown in Figure~\ref{fig1: example}, the samples with conflicting labels in their neighbors should expect high dissonance.  These samples are often close to the classification boundary. In addition, outliers (or out-of-distribution samples) should expect high vacuity. 

Let $\mathcal{D}_{\text{OOD}}$ be the set of out-of-distribution samples in $\mathcal{D}$ and $\mathcal{D}_{\text{BOD}}$ be the set of samples whose neighboring samples have conflicting evidence, where the set of samples in in-distribution ($\mathcal{D}_{\text{IN}})$ is set to $\mathcal{D}_{\text{IN}} = \mathcal{D} - \mathcal{D}_{\text{OOD}}- \mathcal{D}_{\text{BOD}}$. We then propose a training method using a regularized ENN that minimizes the following loss function:
\begin{eqnarray}
\mathcal{L}(\Theta) = \mathbb{E}_{({\bf x}_i, {\bf y}_i) \sim \mathcal{D}}[\mathcal{L}(f({\bf x}_i|\Theta), {\bf y}_i) ] -\nonumber \\
 \lambda_1   \mathbb{E}_{({\bf x}_i, {\bf y}_i) \sim \mathcal{D}_{\text{OOD}}}[\text{Vac}(f({\bf x}_i|\Theta))] - \nonumber \\
 \lambda_2   \mathbb{E}_{({\bf x}_i, {\bf y}_i) \sim \mathcal{D}_{\text{BOD}}}[\text{Diss}(f({\bf x}_i|\Theta))],
\end{eqnarray}
over the parameters $\Theta$ of $f$, where $\lambda_1$ and $\lambda_2$ are the tradeoff parameters that decide the importance of the expected uncertainty dimensions associated with vacuity and dissonance. 

\noindent {\bf KL divergence}:for each training point ${\bf x}_i$, we find $K$ nearest neighbors, and consider their class labels as the class observations of ${\bf x}_i$, in this way, we will be able to estimate the Dirichlet parameters $\hat{\alpha}_i$ for the training point ${\bf x}_i$. We should learn a ENN such that the KL divergence between predicted Dirichlet parameters $\alpha_i$ and the estimated Dirichlet parameters $\hat{\alpha}_i$ based on $K$ nearest neighbors for each training point should be small, the new objective:
\begin{eqnarray}
\mathcal{L}(\Theta) + \sum_{{\bf x}_i\in \mathcal{D}}\text{KL}[\text{Dir}(\mathbf{p}_i;{\bm \alpha}_i)||\text{Dir}(\hat{\mathbf{p}_i};\hat{\alpha}_i)]
\end{eqnarray}

\section{Experiments \& Results} \label{sec:results}
We evaluate our proposed method following the experimental setup in~\cite{louizos2017multiplicative}. We use the standard LeNet with ReLU non-linearities as the neural network architecture. All experiments are implemented in Tensorflow~\cite{tensorflow2015-whitepaper} where the Adam optimizer~\cite{kingma2014adam} is used with default settings for training\footnote{For review purpose, the source code and datasets are accessible at \url{https://www.dropbox.com/sh/uhonftulu9x2xa9/AABZxzeraWN8SYHh9N_10QoTa?dl=0}}.

\vspace{-2mm}
\subsection{Experimental Setup}
\vspace{-1mm}

\vspace{1mm}
\noindent {\bf Dataset}: The CIFAR10 dataset consists of 60,000 $32 \times 32$ colour images in 10 classes, with 6,000 images per class. The original classification setting used 50,000 training images and 10,000 test images. 

\vspace{1mm}
\noindent {\bf Out-of-Distribution (OOD) Setting}: We used a subset of the classes in the CIFAR10 dataset for training and the rest of the subset for testing OOD uncertainty. To be specific, for training, we use the samples from the first five categories \{airplane, automobile, bird, cat, deer\} in the training set from the CIFAR10 dataset as $\mathcal{D}_{\text{IN}}$ (i.e., in-distribution), another two categories \{ship, truck\} as $\mathcal{D}_{\text{OD}}$ (i.e., out-of-distribution), and the left three categories \{dog, frog, horse\} for testing the OOD. For $\mathcal{D}_{\text{BOD}}$ (i.e., boundary-out-of-distribution) dataset, we select the boundary sample by taking the following steps: (1) calculate the distance between each training image; (2) for each image, choose top 10 similar images as observed evidence; (3) calculate dissonance based the 10 pieces of observed evidence for each training image; and (4) choose the 500 images with the highest dissonance. Here we measure the similarity based on the cosine similarity metric~\cite{bishop2006PRML}. by:
\begin{eqnarray}
\mathrm{Sim} (A, B) = \frac{A\cdot B}{\|A\|_2\cdot \|B\|_2},
\end{eqnarray}
where $A$ and $B$ are the feature vectors  of two images and the raw image pixels are treated as the features for the cosine similarity.

\vspace{1mm}
\noindent {\bf Synthetic Dataset}: We generate a dataset of three classes (1, 2, and 3) in a two dimensional space. We sample 1,000 points of each class (1, 2, and 3) from $\mathcal{N}\Big(
\begin{pmatrix}
-2  \\
-2 \\
\end{pmatrix}
, \begin{pmatrix}
1 & 0  \\
0 & 1 \\
\end{pmatrix}\Big)$, $\mathcal{N}\Big(
\begin{pmatrix}
0  \\
1.464 \\
\end{pmatrix}
, \begin{pmatrix}
1 & 0  \\
0 & 1 \\
\end{pmatrix}\Big)$, and $\mathcal{N}\Big(
\begin{pmatrix}
2  \\
-2 \\
\end{pmatrix}
, \begin{pmatrix}
1 & 0  \\
0 & 1 \\
\end{pmatrix}\Big)$, respectively.  In addition, we sample 100 OOD points from $\mathcal{N}\Big(
\begin{pmatrix}
-8  \\
-8 \\
\end{pmatrix}
, \begin{pmatrix}
1 & 0  \\
0 & 1 \\
\end{pmatrix}\Big)$ and $\mathcal{N}\Big(
\begin{pmatrix}
8 \\
-8 \\
\end{pmatrix}
, \begin{pmatrix}
1 & 0  \\
0 & 1 \\
\end{pmatrix}\Big)$ each. Therefore, we have 3,000 in-distribution points and 200 out-of-distribution points in total. We expect that the testing points near the boundary of the three classes should expect high dissonance measures and the testing points that are far away from the 3,000 training points should expect high vacuity measures.

\vspace{1mm}
\noindent {\bf Comparison Schemes}: We compared the following approaches: 
\begin{itemize}
\item \textbf{L2} corresponds to the standard deterministic neural nets with softmax output and weight decay;
\item \textbf{ENN} refers to evidential deep learning model~\cite{sensoy2018evidential}; 
\item \textbf{ENN-Vac} uses ENN with the vacuity regularization only; 
\item \textbf{ENN-Diss} uses ENN with the dissonance regularization only;
\item \textbf{ENN-Vac-Diss} uses ENN with both the vacuity and dissonance regularization.

\end{itemize}

\vspace{1mm}
\noindent {\bf Parameter Tuning}: We trained the same model LeNet architecture using 20 and 50 filters with size $5 \times 5$ at the first and second convolutional layers, respectively; and 500 hidden units are used for the fully connected layer, same as in the ENN model~\cite{sensoy2018evidential}. For all comparison methods, we set $\mathrm{batch\_size}=1000$, $\mathrm{learning\_rate}=0.01$, $\mathrm{dropout\_rate}=0.9$ and $\mathrm{weight\_decay\_coefficient} = 0.005$. For vacuity and dissonance regularization parameters, we set $\lambda_1 = 0.01, \lambda_2=0.01$.  The values of the parameters used are obtained by tuning them to show the optimal effect of using regularization for synthetic data.

\vspace{1mm}
\noindent {\bf Metrics}: Our approach directly quantifies vacuity and dissonance based on Eqs.~\eqref{vacuity} and~\eqref{eq:belief-dissonance-multi}, respectively. However, some approaches use entropy to measure prediction uncertainty as in~\cite{louizos2017multiplicative}, i.e., uncertainty in predicted probabilities increases as their entropy increases. To be fair, we use the same metric for evaluating prediction uncertainty for OOD in the rest of the paper and for class probabilities. We use the empirical CDFs over the range of possible entropy $[0, \log 5]$ for all models trained with the CIFAR10 dataset. The curves closer to the bottom right corner of the plot are considered desirable, indicating maximum entropy in all predictions, leading to high OOD detection~\cite{louizos2017multiplicative}.

\begin{figure*}[t]
\centering
\begin{subfigure}{.49\textwidth}
  \centering
  \includegraphics[width=.834\linewidth, height=.574\linewidth]{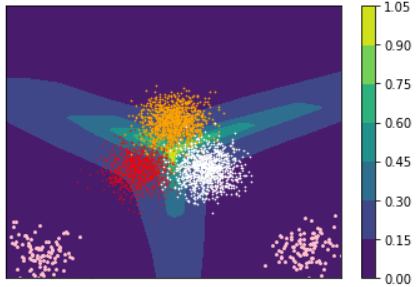}
  \caption{Vacuity contour (ENN model)}
  \label{fig2:synthetic a}
\end{subfigure}
\begin{subfigure}{.49\textwidth}
  \centering
  \includegraphics[width=.834\linewidth, height=.574\linewidth]{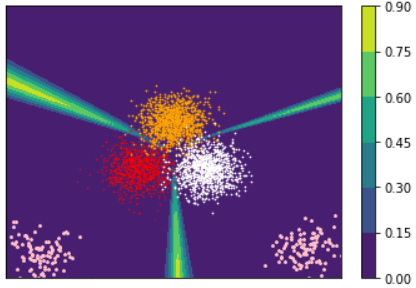}
  \caption{Dissonance contour map (ENN model)}
  \label{fig2:synthetic b}
\end{subfigure}
\begin{subfigure}{.49\textwidth}
  \centering
  \includegraphics[width=.834\linewidth, height=.574\linewidth]{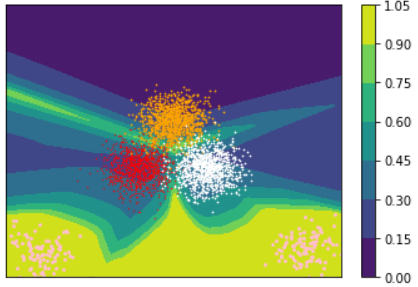}
  \caption{Vacuity contour map (ENN-Vac model)}
  \label{fig2:synthetic c}
\end{subfigure}
\begin{subfigure}{.49\textwidth}
  \centering
  \includegraphics[width=.834\linewidth, height=.574\linewidth]{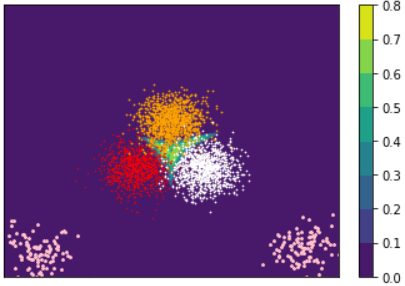}
  \caption{Dissonance contour map (ENN-Vac-Diss model)} \label{fig2:synthetic d}
\end{subfigure}
\caption{Comparison between the vacuity and dissonance of the predictions of ENN ((a) and (b)) and regularized ENN ((c) and (d)) where each node's feature value is represented by a coordinate with the range in $[-10,10] \times[-10,10]$.}
\label{fig2:synthetic}
\end{figure*}

\begin{figure}[ht]
    \centering
    \includegraphics[width=1.0\linewidth]{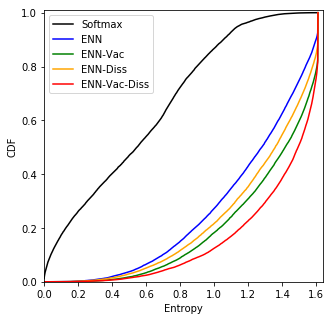}
    \caption{Empirical CDF for the entropy of the predictive distributions on the subset of the classes of the CIFAR10 dataset.}
    \label{fig:ood-cifar}
\end{figure}

\subsection{Experimental Results}
{\bf Results with Synthetic Datasets}: In this experiment, we trained the model based on the in-distribution class (i.e., yellow, red and white nodes) and tested the rest of nodes except that ENN-Vac and ENN-Vac-Diss also use outlier nodes (pink nodes at both bottom left and bottom right corners) in the training, ranging in $[-10,10]\times[-10,10]$, where each coordinate represents the values of node features in the given synthetic dataset. In Figure~\ref{fig2:synthetic}, the yellow, red and white nodes are training nodes where the contour map indicates the different level of strengths from purple (weak, low uncertainty) to green (strong, high uncertainty) for a given uncertainty (i.e., vacuity, dissonance). Here, we evaluate the performance of the all four schemes (i.e., ENN, ENN-Vac, ENN-Diss, and ENN-Vac-Diss) in terms of classification predictions and how different uncertainty types are measured under different schemes. Figures~\ref{fig2:synthetic a} and~\ref{fig2:synthetic c} show vacuity from ENN and ENN-Vac, respectively. The observed high vacuity in Figure~\ref{fig2:synthetic c} (ENN-Vac) is more reasonable as the opinions of nodes are far away from the training data.  That is, since OOD class nodes are expected to have lack of evidence with high vacuity. Region influenced by our OOD training nodes extend to the whole bottom area which gives more advantages of OOD detection.  Figures~\ref{fig2:synthetic b} and~\ref{fig2:synthetic d} show the observed dissonance of nodes' subjective opinions. As expected, Figure~\ref{fig2:synthetic d} shows a more reasonable level of dissonance than Figure~\ref{fig2:synthetic b}, exhibiting stronger dissonance in the boundary areas among three classes. Figure~\ref{fig2:synthetic} also demonstrates that the choice of OOD samples is critical for Vacuity to be high far from the training data. Ideally, the vacuity should be high at the upper portions and sides of Figure~\ref{fig2:synthetic c} as well. It indicates that the selection of appropriate OOD samples affects the accurate estimation of vacuity. 

\vspace{1mm}
\noindent {\bf Results with Cifar Datasets}:  Figure~\ref{fig:ood-cifar} shows the empirical CDFs for all models trained with the CIFAR10 dataset. The curves closer to the bottom right corner of the plot are desirable, indicating maximum entropy in all predictions~\cite{louizos2017multiplicative}. It is clear that the uncertainty estimates of our model ENN-Vac is significantly better than those of the baseline methods (ENN and Softmax), which shows the benefit of using vacuity as an indicator of uncertainty to improve OOD detection. ENN-Diss is comparable with ENN, showing that dissonance regularization does not help significantly for the OOD detection. One interesting pattern shows that ENN-Vac-Diss outperformed among all other methods, demonstrating that considering both vacuity and dissonance regularized can significantly improve the OOD detection.

\section{Conclusion \& Future Work} \label{sec:conclusion-future-work}
In this work, we developed the regularized ENNs that learn an ENN based on the regularization associated with two different uncertainty dimensions, vacuity and dissonance.  Via extensive experiments based on both synthetic and real datasets, we demonstrated more reasonable levels of estimated vacuity and dissonance in classification prediction and out-of-distribution detection. From our study, the following {\bf key findings} are obtained:
\begin{itemize}
\item Overall our proposed ENN considering both vacuity and dissonance (i.e., ENN-Vac-Diss) outperformed among all in terms of the performance in the out-of-distribution task. In addition, our proposed ENN with vacuity considered (ENN-Vac) outperformed among all baselines as the vacuity is regularized in an ENN.
\item Our proposed ENN-Vac and ENN using dissonance (ENN-Diss) showed more desirable uncertainty measurement for vacuity and dissonance, respectively.  To be specific, ENN-Vac showed high vacuity in more outliers (out-of-distribution) samples. In addition, when ENN-Diss is used, higher dissonance is observed particularly at the boundary areas between different classes.  This provides great promise for this regularized ENNs considering different uncertainty dimensions to be applicable in various types of classification prediction and OOD detection tasks.   
\end{itemize}

In our future work, we plan to: (1) investigate the clear reasons of unstable performance of ENN considering multiple dimensions of uncertainty; (2) investigate  how OOD samples can be generated to more effectively estimate vacuity; (3) validate the performance of our \textit{regularized} ENNs approach based on more real-world datasets (e.g., object detection datasets); and (4) extend our proposed work with more uncertainty regularized that can deal with vagueness in hyper opinions in Subjective Logic.

\bibliographystyle{aaai}
\bibliography{main}

\end{document}